\documentclass[]{article}
\usepackage{epsfig, fullpage, amsmath}

\input epsf

\begin{document}

\newcommand{\dima}[1]{{{#1}}}
\newcommand{\zchaff}{{\tt zChaff}}
\newcommand{\satz}{{\tt Satz}}
\newcommand{\walksat}{{\tt WalkSAT}}
\newcommand{\survey}{{\tt SP}}
\newcommand{\uc}{{\tt UC}}
\newcommand{\whp}{w.h.p.}
\newcommand{\ie}{i.e.,\ }
\newcommand{\ex}{{\bf E}}
\def\eg{e.g.\ }
\newcommand{\ds}{{\rm d}s}
\newcommand{\dc}{{\rm d}c}
\newcommand{\dr}{{\rm d}r}
\newcommand{\dx}{{\rm d}x}

\newcommand{\remove}[1]{}

\title{Hiding Satisfying Assignments: Two are Better than One}

\author{\begin{tabular}[t]{c@{\extracolsep{1em}}c@{\extracolsep{1em}}c }
Dimitris Achlioptas & Haixia Jia & Cristopher Moore\\
optas@microsoft.com & hjia@cs.unm.edu  & moore@cs.unm.edu \\
Microsoft Research  & Computer Science Department &Computer Science Department\\
Redmond, Washington & University of New Mexico & University of New Mexico
\end{tabular}}

\date{}


\maketitle
\begin{abstract}
The evaluation of incomplete satisfiability solvers depends
critically on the availability of hard satisfiable instances. A
plausible source of such instances consists of random $k$-SAT
formulas whose clauses are chosen uniformly from among all clauses
satisfying some randomly chosen truth assignment $A$.
Unfortunately, instances generated in this manner tend to be
relatively easy and can be solved efficiently by practical
heuristics. Roughly speaking, as the formula's density increases,
for a number of different algorithms, $A$ acts as a stronger and
stronger attractor. Motivated by recent results on the geometry of
the space of satisfying truth assignments of random $k$-SAT and
NAE-$k$-SAT formulas, we introduce a simple twist on this basic
model, which appears to dramatically increase its hardness.
Namely, in addition to forbidding the clauses violated by the
hidden assignment $A$, we also forbid the clauses violated by its
complement, so that both $A$ and $\overline{A}$ are satisfying. It
appears that under this ``symmetrization'' the effects of the two
attractors largely cancel out, making it much harder for
algorithms to find any truth assignment. We give theoretical and
experimental evidence supporting this assertion.
\end{abstract}

\section{Introduction}

Recent years have witnessed the rapid development and application
of search methods for constraint satisfaction and Boolean
satisfiability. An important factor in the success of these
algorithms is the availability of good sets of benchmark problems
to evaluate and fine-tune them. There are two main sources of such
problems: the real world, and random instance generators.
Real-world problems are arguably the best benchmark, but
unfortunately are often in short supply. Moreover, using real-world
problems carries the risk of tuning algorithms toward the specific
application domains for which good benchmarks are available. In
that sense, random instance generators are a good additional
source, with the advantage of controllable characteristics, such
as size and expected hardness.

Hard random instances have led to the development of new
stochastic search methods such as \walksat\ \cite{Selman96} and
the breakout procedure~\cite{Morris}, and have been used in
detailed comparisons of local search methods for graph coloring
and related graph problems~\cite{Johnson}.  The results of various
competitions for CSP and SAT algorithms show a fairly direct
correlation between the performance on real-world benchmarks and
on hard random instances~\cite{dimacs93,dimacs96,Johnson}.
Nevertheless, a key limitation of current problem generators
concerns their use in evaluating {\em incomplete\/} satisfiability
solvers such as those based on local search methods.

When an incomplete algorithm does not find a solution, it can be
difficult to determine whether this is because the instance is in
fact unsatisfiable, or simply because the algorithm failed to find
the satisfying assignment. The standard way of dealing with this
problem is to use a complete search method to filter out the
unsatisfiable cases. However, this greatly limits the size and
difficulty of problem instances that can be considered. Ideally,
one would use problem generators that generate satisfiable
instances only. One relatively recent source of such problems is
the quasigroup completion
problem~\cite{ssw,quasigroup2,quasigroup1}. However, a generator
for random hard satisfiable instances of 3-SAT, say, has remained
elusive.

Perhaps the most natural candidate for generating random hard
satisfiable 3-SAT formulas is the following. Pick a random truth
assignment $A$, and then generate a formula with $n$ variables and
$rn$ random clauses, rejecting any clause that is violated by $A$.
In particular, we might hope that if we work close to the
satisfiability threshold region $r \approx 4.25$, where the
hardest random 3-SAT problems seem to be~\cite{cheeseman,hh,msl},
this would generate hard satisfiable instances. Unfortunately,
this generator is highly biased towards formulas with many
assignments clustered around $A$. When given to local search
methods such as \walksat, the resulting formulas turn out to be
much easier than formulas of comparable size obtained by filtering
satisfiable instances from a 3-SAT generator. More sophisticated
versions of this ``hidden assignment''
scheme~\cite{Asahiro,VanGelder} improve matters somewhat but still
lead to easily solvable formulas.

In this paper we introduce a new generator of random satisfiable
problems. The idea is simple: we pick a random 3-SAT formula that
has a ``hidden'' {\bf complementary pair} of satisfying
assignments, $A$ and $\overline{A}$, by rejecting clauses that are
violated by either $A$ or $\overline{A}$. We call these
``2-hidden'' formulas. Our motivation comes from recent
work~\cite{achmooresicomp,achmooreksat} which showed that moving from random
$k$-SAT to random NAE-$k$-SAT (in which every clause in the
formula must have at least one true {\em and\/} at least one false
literal) tremendously reduces the correlation between solutions.
That is, whereas in random $k$-SAT, satisfying assignments tend to
form clumps, in random NAE-$k$-SAT the solutions appear to be
scattered throughout $\{0,1\}^n$ in a rather uniform ``mist'',
even for densities extremely close to the threshold. An intuitive
explanation for this phenomenon is that since the complement of
every NAE-assignment is also an NAE-assignment, the attractions of
solution pairs largely ``cancel out.''  In this paper we exploit
this phenomenon to impose a similar symmetry on the hidden
assignments $A$ and $\overline{A}$, so that {\em their\/}
attractions cancel out, making it hard for a wide variety of
algorithms to ``feel'' either one.

A particularly nice feature of our generator is that it is based
on an extremely simple probabilistic procedure, in sharp contrast
with 3-SAT generators based on, say, cryptographic
ideas~\cite{crypto}. In particular, our generator is readily
amenable to all the mathematical tools that have been developed
for the rigorous  study of random $k$-SAT formulas. Here we make
two first steps in that direction. In Section~\ref{sec:structure},
via a first moment calculation we study the distribution of the
number of solutions as a function of their distance from the
hidden assignments. In Section~\ref{sec:uc}, on the other hand, we
use the technique of differential equations to analyze the
performance of the Unit Clause (\uc) heuristic on our formulas.

Naturally, mathematical simplicity would not be worth much if the
formulas produced by our generator were easily solvable. In
Section~\ref{sec:exp}, we compare experimentally the hardness of
``2-hidden'' formulas with that of ``1-hidden'' and ``0-hidden''
formulas. That is, we compare our formulas with random 3-SAT
formulas with one hidden assignment and with standard random 3-SAT
formulas with no hidden assignment. We examine four leading
algorithms: two complete solvers, \zchaff\ and \satz, and two
incomplete ones, \walksat\ and the recently introduced Survey
Propagation (\survey).

For all these algorithms, we find that our formulas are much
harder than 1-hidden formulas and, more importantly, {\em about as
hard as 0-hidden formulas,} of the same size and density.

\section{A picture of the space of solutions}
\label{sec:structure}

In this section we compare 1-hidden and 2-hidden formulas with
respect to the expected number of solutions at a given distance
from the hidden assignment(s).

\subsection{1-hidden formulas}

Let $X$ be the number of satisfying truth assignments in a random
$k$-SAT formula with $n$ variables and $m=rn$ clauses chosen
uniformly and independently among all $k$-clauses with {\em at
least one positive literal}, \ie 1-hidden formulas where we hide
the all--ones truth assignment.  To calculate the expectation
$\ex[X]$, it is helpful to parametrize truth assignments according
to their {\em overlap} with the hidden assignment, i.e., the
fraction $\alpha$ of variables on which they agree with $A$, which
in this case is the fraction of variables that are set to one.
Then, linearity of expectation gives~\eqref{linear}, clause
independence gives~\eqref{clause}, selecting the literals in each
clause uniformly and independently gives~\eqref{replacement}, and,
finally, writing $z=\alpha n$ and using Stirling's approximation
for the factorial gives~\eqref{stirli}:
\begin{eqnarray}
\ex[X] & = & \sum_{A \in \{0,1\}^n} \Pr[A \mbox{ is satisfying}] \label{linear} \\
& = & \sum_{z=0}^n \binom{n}{z} \Pr[\mbox{a truth assignment with
$z$ ones satisfies a random clause}]^m \label{clause}\\
& = & \sum_{z=0}^n \binom{n}{z} \left(1-\frac{1}{2^k-1}\sum_{j=1}^k\binom{k}{j}(1-z/n)^{j}(z/n)^{k-j}\right)^m\label{replacement}\\
& = & \sum_{z=0}^n \binom{n}{z} \left(1-\frac{1-(z/n)^k}{2^k-1}\right)^m \nonumber\\
& = & {\mathrm{poly}}(n) \times \max_{\alpha \in [0,1]}\left[
\frac{1}{\alpha^{\alpha}(1-\alpha)^{1-\alpha}}\left(1-\frac{1-\alpha^k}{2^k-1}\right)^r\right]^n\label{stirli}\\
 & \equiv & {\mathrm{poly}}(n) \times
\max_{\alpha \in [0,1]}
[f_{k,r}(\alpha)]^{n^{\phantom{k^{k^{k}}}}} \nonumber
 \end{eqnarray}
where
\[ f_{k,r}(\alpha) = \frac{1}{\alpha^{\alpha}(1-\alpha)^{1-\alpha}}\left(1-\frac{1-\alpha^k}{2^k-1}\right)^r
\enspace . \] From this calculation we see that $\ex[X]$ is
dominated by the contribution of the truth assignments that
maximize $f_{k,r}(\alpha)$ (since we raise $f_{k,r}$ to the $n$th
power all other contributions vanish). Now, we readily see that
$f$ is the product of an ``entropic'' factor $1/(\alpha^\alpha
(1-\alpha)^{1-\alpha})$ which is symmetric around $\alpha=1/2$,
and a ``correlation'' factor which is strictly increasing in
$\alpha$. As a result, it is always maximized for some
$\alpha>1/2$. This means that the dominant contribution to
$\ex[X]$ comes from truth assignments that agree with the hidden
assignment on more that half the variables. That is, the set of
solutions is dominated by truth assignments that can ``feel'' the
hidden assignments. Moreover, as $r$ increases this phenomenon
becomes more and more acute (see Figure~\ref{opala} below).

\subsection{2-hidden formulas}

Now let $X$ be the number of satisfying truth assignments in a
random $k$-SAT formula with $n$ variables and $m=rn$ clauses
chosen uniformly among all $k$-clauses that have at least one
positive {\em and at least one negative literal}, \ie 2-hidden
formulas where we hide the all--ones assignment {\em and\/} its
complement. To compute $\ex[X]$ we proceed as above, except that
now~\eqref{replacement} is replaced by
\[
\sum_{z=0}^n \binom{n}{z}
\left(1-\frac{1}{2^k-2}\sum_{j=1}^{k-1}\binom{k}{j}(1-z/n)^{j}(z/n)^{k-j}\right)^m
\enspace .
\]
Carrying through the ensuing changes we find that now
\[ \ex[X] = {\mathrm{poly}}(n) \times
\max_{\alpha\in[0,1]}[g_{k,r}(\alpha)]^n \]
where
\[
g_{k,r}(\alpha)
=\frac{1}{\alpha^{\alpha}(1-\alpha)^{1-\alpha}}\left(1-\frac{1-\alpha^k
- (1-\alpha)^k}{2^k-2}\right)^r \enspace .
\]
This time, both the entropic factor and the correlation factor
comprising $g$ are symmetric functions of $\alpha$, so $g_{k,r}$
is symmetric around $\alpha=1/2$ (unlike $f_{k,r}$). Indeed, one
can prove that for all $r$ up to {\em extremely close\/} to the
random $k$-SAT threshold $r_k$, the function $g_{k,r}$ has its
global maximum at $\alpha = 1/2$. In other words, for all such
$r$, the dominant contribution to $\ex[X]$ comes from truth
assignments at distance $n/2$ from the hidden assignments, \ie the
hidden assignments are ``not felt.''  More precisely, there exists
a sequence $\epsilon_k \rightarrow 0$ such that $g_{k,r}$ has a
unique global maximum at $\alpha=1/2$, for all
\begin{equation}\label{lower}
 r\leq 2^k \ln 2 - \frac{\ln 2}{2} -1 - \epsilon_k \enspace .
\end{equation}

Contrast this with the fact (implicit in~\cite{KKKS}) that  for
\begin{equation}\label{upper}
r \geq 2^k \ln 2 - \frac{\ln 2}{2} - \frac{1}{2} \enspace ,
\end{equation}
a random $k$-SAT formula with $n$ variables and $m=rn$ clauses is
unsatisfiable with probability $1-o(1)$. Moreover, the convergence
of the sequence $\epsilon_k \to 0$ is rapid, as can be seen from
the concrete values in table~\ref{table:val}.
\begin{table}[h]
$$
\begin{array}{l|cccccc}
 \hspace*{0.1cm} k              &  \makebox[1.2cm]{3}   & \makebox[1.3cm]{4}  & \makebox[1.4cm]{5}         &  \makebox[1.5cm]{7}        &  \makebox[1.6cm]{10}       &   \makebox[1.7cm]{20} \\   \hline
        \mbox{Eq.\ \eqref{lower}}        &   7/2     &   35/4    &   20.38   &   87.23   &   708.40  & 726816.15 \\
        \mbox{Eq.\ \eqref{upper}}        &   4.67    &   10.23   &   21.33   &   87.88   &   708.94  & 726816.66
\end{array}
$$
\caption{The convergence (in $k$) to the asymptotic gap of $1/2$ is rapid}
\label{table:val}
\end{table}

Below we plot $f_{k,r}$ and $g_{k,r}$ for $k=5$ and
$r=16,18,20,22,24$ (from top to bottom). We see that in the case
of 1-hidden formulas, \ie $f_{k,r}$, the maximum always occurs to
the right of $\alpha=1/2$. Moreover, observe that for $r=22,24$,
\ie after we cross the 5-SAT threshold (which occurs at $r \approx
21$) we have a dramatic shift in the location of the maximum and,
thus, in the extent of the bias: as one would expect, the only
remaining satisfying assignments above the threshold are those
extremely close to the hidden assignment.

In the case of 2-hidden formulas, on the other hand, we see
that for $r=16,18,20$ the global maximum occurs at $\alpha=1/2$
(from the table above we know that the critical $r$ for $k=5$ is
$20.38$). For $r=20$, we also have two local maxima, near
$\alpha=0,1$, but since $g_{k,r}$ is raised to the $n$th power,
these are exponentially suppressed. Naturally, for $r$ above the
threshold, \ie $r=22,24$, these local maxima become global,
signifying that indeed the only remaining truth assignments are
those extremely close to one of the two hidden ones.

Intuitively, we expect that because $g$ is flat at $\alpha = 1/2$
where random truth assignments are concentrated, for 2-hidden formulas
local search algorithms like \walksat\ will essentially perform a
random walk until they are lucky enough to get close to one of the
two hidden assignments.  Thus we expect \walksat\ to take about as long
on 2-hidden formulas as it does on 0-hidden ones.
For 1-hidden formulas, in contrast, we expect the nonzero gradient
of $f$ at $\alpha = 1/2$ to provide a strong ``hint'' to \walksat\
that it should move towards the hidden assignment, and that therefore
1-hidden formulas will be much easier for it to solve.  We will see below
that our experimental results bear out these intuitions perfectly.

\begin{figure}\label{fig:dist}
\begin{center}
\begin{minipage}{3in}
\begin{center}
        \centerline{\hbox{
        \psfig{figure=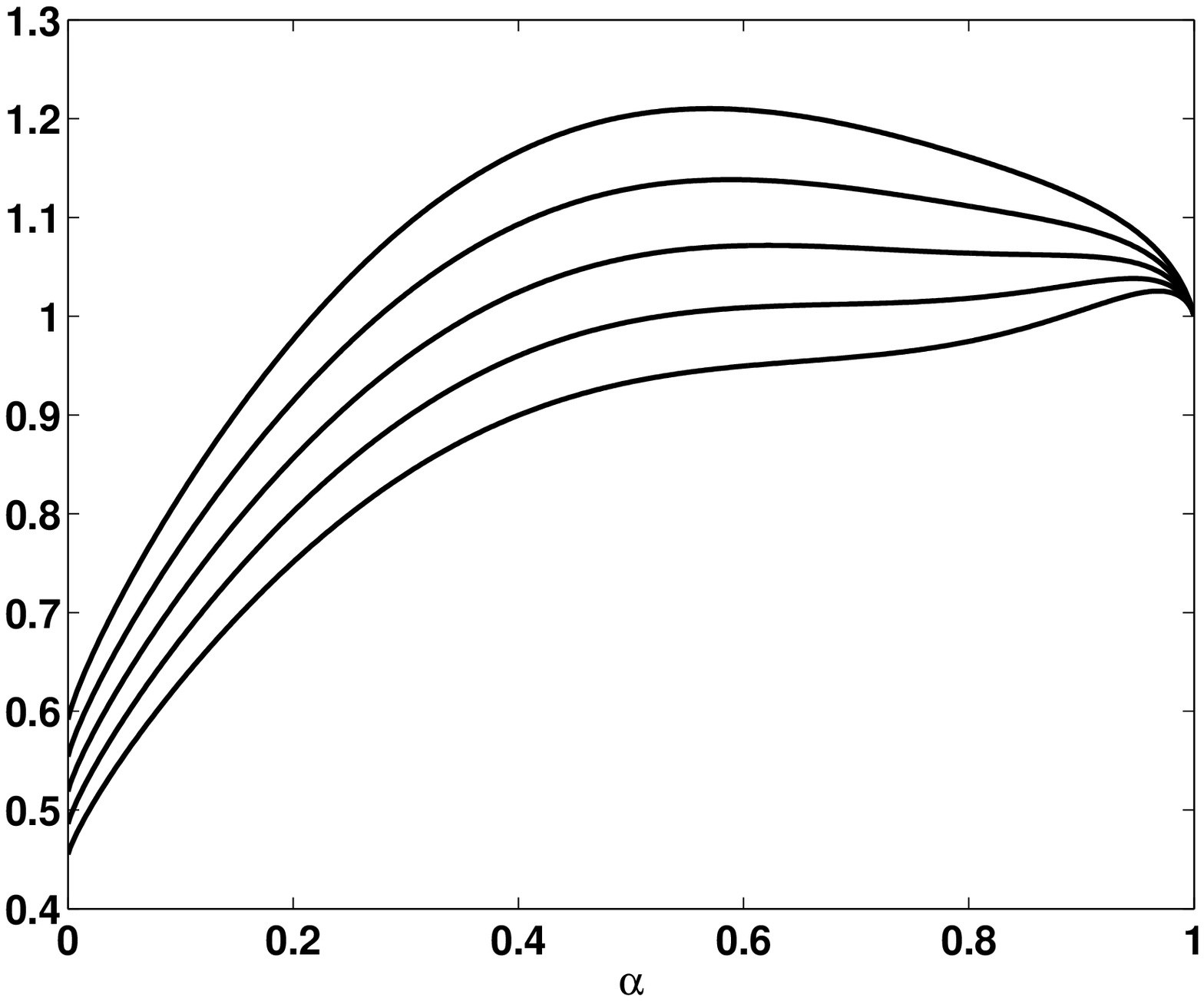,height=1.7in,width=2.5in}
        }}
        \centerline{1-hidden formulas}
\end{center}
\end{minipage}\    \
\begin{minipage}{3in}
\begin{center}
        \centerline{\hbox{
        \psfig{figure=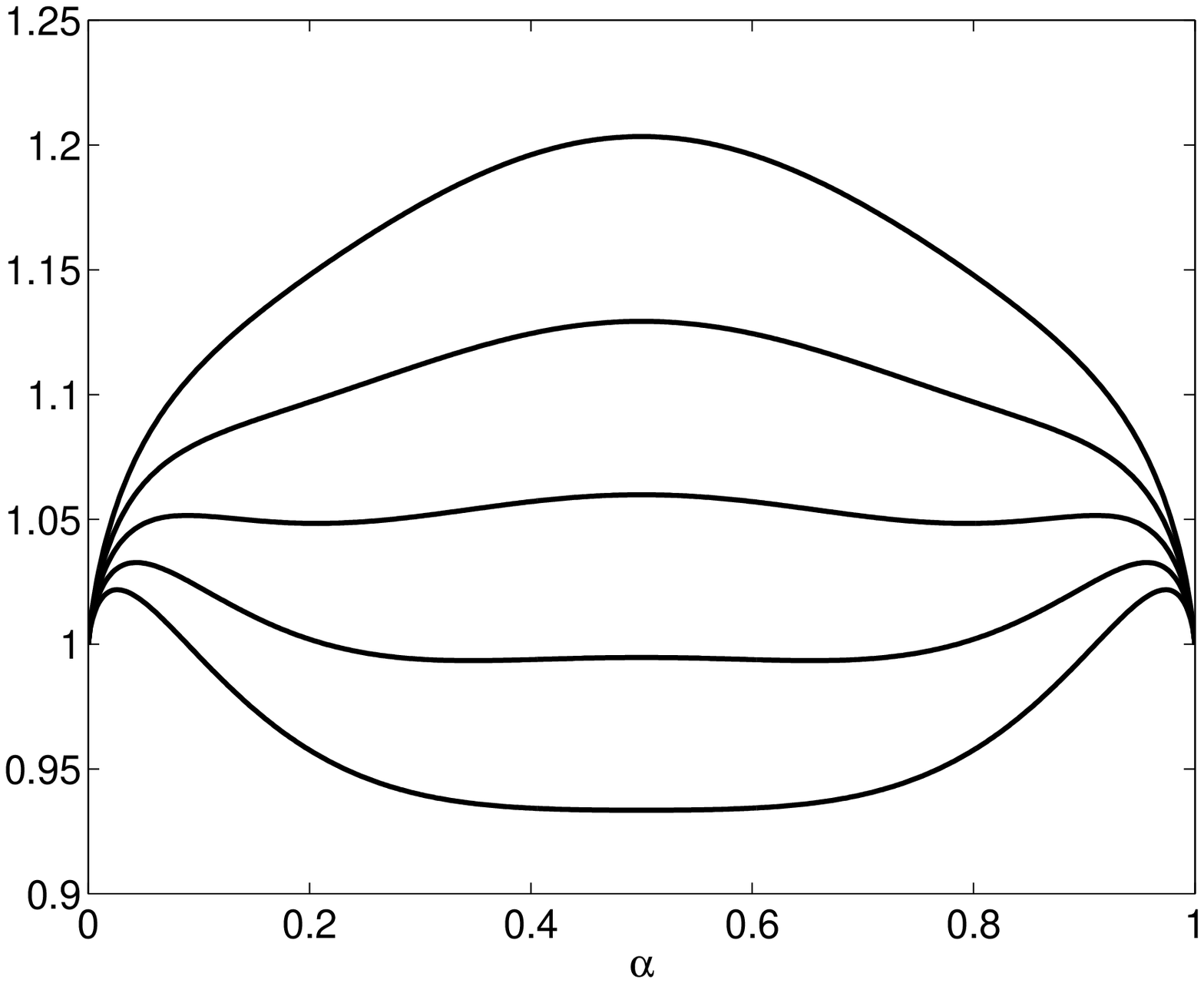,height=1.7in,width=2.5in}
        }}
        \centerline{2-hidden formulas}
\end{center}
\end{minipage}
\end{center}

\vspace*{-.6cm}

\caption{The $n$th root of the expected number of solutions $f_{k,r}$ and $g_{k,r}$ for 1-hidden and 2-hidden formulas respectively, as a function of the overlap fraction $\alpha=z/n$ with the hidden assignment. Here $k=5$ and $r=16, 18, 20, 22, 24$ from top to bottom.}\label{opala}
\end{figure}

\section{The Unit Clause heuristic and DPLL algorithms}
\label{sec:uc}

Consider the following linear-time heuristic, called Unit Clause
(\uc), which permanently sets one variable in each step as
follows: pick a random literal and satisfy it; repeatedly satisfy
any 1-clauses present. In~\cite{chaofranco}, Chao and Franco
showed that \uc\ succeeds with constant probability on random
3-SAT formulas with $r < 8/3$, and fails with high
probability, i.e., with probability $1-o(1)$ as $n \to \infty$, 
for $r > 8/3$. One can think of \uc\ as the first branch of the simplest
possible DPLL algorithm $S$: set variables in a random order, each
time choosing randomly which branch to take first.
The result of~\cite{chaofranco} then shows that, with constant probability,
$S$ solves random 3-SAT formulas with $r < 8/3$ with no backtracking
at all.

Conversely, calculations from statistical
physics~\cite{monasson1,monasson2} suggest that with high
probability $S$ takes exponential time for all $r > 8/3$. That is,
around $r=8/3$ the running time of $S$ goes from linear to
exponential, with no intermediate regime. In~\cite{abm}, it was
proved that $S$ takes exponential time for $r > 3.81$, which is
already well below the conjectured satisfiability threshold $r
\approx 4.2$. Moreover, the results in~\cite{abm} imply that if
the ``tricritical point'' of $(2+p)$-SAT is $r=2/5$, one can
replace 3.81 with 8/3.

In this section we analyze the performance of
\uc\ on 1-hidden and 2-hidden formulas.  Specifically, we show that \uc\ fails
for 2-hidden formulas at precisely the same density as for 0-hidden ones.
Based on this, we conjecture that the running time of $S$, and other
simple DPLL algorithms, becomes exponential for 2-hidden formulas
at the same density as for 0-hidden ones.

\remove{
A key observation in~\cite{abm} is that if a DPLL algorithm uses
a ``myopic'' splitting rule---that is, one which chooses the residual formula
remaining before the first backtrack of a DPLL algorithm is
uniformly random, conditional on the number of clauses of each
length. Thus, extending the result of Chv{\'a}tal and
Szemer{\'e}di to $(2+p)$-SAT formulas implies that in that first
path, \whp\ a myopic algorithm creates a residual formula which is
unsatisfiable and which only has exponentially long resolution
refutations. Our analysis for \uc\ below shows that the
``action'' of this heuristic on a random 2-hidden formula is
identical to its action on a random 0-hidden formula (and unlike
its action on a 1-hidden formula). This suggests the possibility
that 2-hidden formulas are, in fact, {\em provably\/} hard for
myopic DPLL algorithms, causing them to take exponential time
at the same densities that 0-hidden formulas do.
}

To analyze \uc\ on random 1-hidden and 2-hidden formulas we
actually analyze \uc\ on arbitrary initial distributions of
3-clauses, \ie where for each $0 \le j \le 3$ we specify the
initial number of 3-clauses with $j$ positive literals and $3-j$
negative ones.  We use the method of differential equations; see
\cite{dimitrisreview} for a review.  To simplify notation, we
assume that $A$ is the all--ones assignment, so that
1-hidden formulas forbid clauses where all
literals are negative, while 2-hidden formulas forbid
all-negative and all-positive clauses.

A {\em round} of \uc\ consists of a free step, in which we
satisfy a random literal, and the
ensuing chain of unit-clause propagations. For
$0 \le i \le 3$ and $0 \le j \le i$, let $S_{i,j} = s_{i,j} n$ be
the number of clauses of length $i$ with $j$ positive literals and
$i-j$ negative ones. We will also refer to the total density of
clauses of size $i$ as $s_i = \sum_j s_{i,j}$.  Let $X=xn$ be the
number of variables set so far.  Our goal is to write the expected
change in these variables in a given round as a function of their
values at the beginning of the round. Note that at the beginning
of each round $S_{1,0} = S_{1,1} = 0$ by definition, so the
``state space'' of our analysis will consist of the variables
$S_{i,j}$ for $i \ge 2$.

It is convenient to define two new quantities, $m_T$ and $m_F$,
which are the expected number of variables set True and False in a
round. We will calculate these below. Then, in terms of $m_T,m_F$,
we have
\begin{eqnarray}
\ex[\Delta S_{3,j}] & = & -(m_T + m_F) \,\frac{3 s_{3,j}}{1-x}
\label{eq:expchange} \\
\ex[\Delta S_{2,j}] & = & -(m_T + m_F) \,\frac{2 s_{2,j}}{1-x}
   + m_F \,\frac{(j+1) s_{3,j+1}}{1-x}
   + m_T \,\frac{(3-j) s_{3,j}}{1-x} \label{eq:expchanger}\\
\ex[\Delta X] & = & -(m_T + m_F) \enspace . \nonumber
\end{eqnarray}
To see this, note that a variable appears positively in a clause of
type $i,j$ with probability $j/(n-X)$, and negatively with
probability $(i-j)/(n-X)$. Thus, the negative terms
in~\eqref{eq:expchange} and~\eqref{eq:expchanger} correspond to
clauses being ``hit'' by the variables set, while the positive
term is the ``flow'' of 3-clauses to 2-clauses.

To calculate $m_T$ and $m_F$, we consider the process by which
unit clauses are created during a round. We can model this with a
two-type branching process, which we analyze as in
\cite{achmoorecoloring}.  Since the free step gives the chosen
variable a random value, we can think of it as creating a unit
clause, which is positive or negative with equal probability. Thus
the initial expected population of unit clauses can be represented
by a vector
\[ p_0 = \left( \begin{array}{c} 1/2 \\ 1/2 \end{array} \right) \]
where the first and second components count
the negative and positive unit clauses respectively.
Moreover, at time $X=xn$, a unit clause procreates
according to the matrix
\[ M = \frac{1}{1-x} \left( \begin{array}{cc}
s_{2,1} & 2 s_{2,0} \\
2 s_{2,2} & s_{2,1}
\end{array} \right) \enspace .
\]
In other words, satisfying a negative unit clause creates, in
expectation, $M_{1,1} = s_{2,1} / (1-x)$ negative unit clauses and
$M_{2,1} = 2 s_{2,2} / (1-x)$ positive unit clauses, and similarly
for satisfying a positive unit clause.

Thus, as long as the largest eigenvalue $\lambda_1$ of $M$ is less
than $1$, the expected number of variables set true or false
during the round is given by
\[ \left( \begin{array}{c} m_F \\ m_T \end{array} \right)
= ( I + M + M^2 + \cdots ) \cdot p_0
= \left( I - M \right)^{-1} \cdot p_0 \]
where $I$ is the identity matrix.
Moreover, as long as $\lambda_1 < 1$ throughout the algorithm, \ie
as long as the branching process is subcritical for all $x$, \uc\
succeeds with constant probability. On the other hand, if
$\lambda_1$ ever exceeds $1$, then the branching process becomes
supercritical, with high probability the unit clauses proliferate
and the algorithm fails. Note that
\begin{equation}
 \lambda_1 = \frac{s_{2,1} + 2 \sqrt{s_{2,0} \,s_{2,2}}}{1-x}
 \enspace .
\label{eq:crit}
\end{equation}
Now let us rescale \eqref{eq:expchange} to give a system of
differential equations for the $s_{i,j}$.  Wormald's
Theorem~\cite{wormald} implies that \whp\ the random variables
$S_{i,j}(xn)$ will be within $o(n)$ of $s_{i,j}(x)\cdot n$  for
all $x$:
\begin{eqnarray}
\frac{\ds_{3,j}}{\dx} & = & -\frac{3 s_{3,j}}{1-x}
\label{eq:rescaled} \\
\frac{\ds_{2,j}}{\dx} & = & -\frac{2 s_{2,j}}{1-x}
   + \frac{m_F}{m_T + m_F} \,\frac{(j+1) s_{3,j+1}}{1-x}
   + \frac{m_T}{m_T + m_F} \,\frac{(3-j) s_{3,j}}{1-x} \nonumber
\end{eqnarray}

Now, suppose our initial distribution of 3-clauses is symmetric,
\ie $s_{3,0}(0) = s_{3,3}(0)$ and $s_{3,1}(0) = s_{3,2}(0)$.  It
is easy to see from \eqref{eq:rescaled} that in that case, both
the 3-clauses and the 2-clauses are symmetric at all times, \ie
$s_{i,j} = s_{i,i-j}$ and $m_F = m_T$.  In that case
$s_{2,1} + 2\sqrt{s_{2,0} s_{2,2}} = s_2$, so the criterion for
subcriticality becomes
\[
 \lambda_1 = \frac{s_2}{1-x} < 1 \enspace .
\]
Moreover, since the system~\eqref{eq:rescaled} is now symmetric with respect
to $j$, summing over $j$ gives the differential equations
\begin{eqnarray*}
\frac{\ds_3}{\dx} & = &  -\frac{3 s_3}{1-x} \\
\frac{\ds_2}{\dx} & = & -\frac{2 s_2}{1-x}
   + \frac{3 s_3}{2(1-x)}
\end{eqnarray*}
which are precisely the differential equations for \uc\ on 0-hidden formulas,
\ie random instances of 3-SAT.

Since 2-hidden formulas correspond to symmetric initial
conditions, we have thus shown that \uc\ succeeds on them with
constant probability if and only if $r < 8/3$, \ie that \uc\ fails
on these formulas at exactly the same density for which it fails
on random 3-SAT instances. (In contrast,
integrating~\eqref{eq:rescaled} with the initial conditions
corresponding to 1-hidden formulas shows that \uc\
succeeds for them at a slightly higher density, up to $r<2.679$.)

Of course, \uc\
can easily be improved by making the free step more intelligent:
for instance,
choosing the variable according the number of its occurrences in the
formula, and using the majority of these occurrences to
decide its truth value.  The best known heuristic of this type
\cite{KKL_talk,HajSorkin} succeeds with constant probability for $r < 3.52$.
However, we believe that much of the progress that has been
made in analyzing the performance of such algorithms
can be ``pushed through'' to 2-hidden formulas.
Specifically, nearly all 
algorithms analyzed so far have the property that given as input a
symmetric initial distribution of 3-clauses, \eg random 3-SAT,
their residual formulas consist of symmetric mixes of 2- and
3-clauses.  As a result, we conjecture that the above methods can
be used to show that such algorithms act on 2-hidden formulas
exactly as they do on 0-hidden ones, failing \whp\ at the same
density.

\remove{
This last point acquires additional significance in the light of
recent results on the onset of backtracking in DPLL algorithms for
random 3-SAT formulas.
}

More generally,  call a DPLL algorithm
{\em myopic} if its splitting rule consists of choosing a random
clause of a given size, based on the current distribution of clause sizes,
and deciding how to satisfy it based on the number of occurrences of its variables
in other clauses.
For a given myopic algorithm $A$, let $r_A$ be the
density below which $A$ succeeds without any backtracking with
constant probability. The results of~\cite{abm} imply the
following statement: if the tricritical point for random
$(2+p)$-SAT is $p_c=2/5$ then {\em every\/} myopic algorithm $A$
takes exponential time for $r > r_A$. Thus, not only \uc, but in
fact a very large class of natural DPLL algorithms, would go from
linear time for $r < r_A$ to exponential time for $r > r_A$. The
fact that the linear-time heuristics corresponding to the first
branch of $A$ act on 2-hidden formulas just as they do on 0-hidden
ones suggests that, for a wide variety of DPLL algorithms,
2-hidden formulas become exponentially hard at the same density as
0-hidden ones.  Proving this, or indeed proving that 2-hidden
formulas take exponential time for $r$ above some critical
density, appears to us a very promising direction for future work.

\section{Experimental results}
\label{sec:exp}

In this section we report experimental results on our 2-hidden
formulas, and compare them to 1-hidden and 0-hidden ones.  We use
two leading complete solvers, \zchaff\ and \satz, and two leading
incomplete solvers, \walksat\ and the new Survey Propagation
algorithm \survey. In an attempt to avoid the numerous spurious
features present in ``too-small'' random instances, \ie in
non-asymptotic behavior, we restricted our attention to
experiments where $n \geq 1000$. This meant that
\zchaff\ and \satz\ could only be examined at densities
significantly above the satisfiability threshold, 
as neither algorithm could practically solve either 0-hidden {\em
or\/} 2-hidden formulas with $n \sim 1000$ variables close to the
threshold. For \walksat\ and \survey, on the other hand, we
can easily run experiments in the hardest range (around the
satisfiability threshold) for $n \sim 10^4$.

\subsection{\zchaff\ and \satz}

In order to do experiments with $n \ge 1000$ with \zchaff\ and
\satz, we focused on the regime where $r$ is relatively large, $20
< r < 60$.  As stated above, for $r$ near the satisfiability
threshold, 0-hidden and 2-hidden random formulas with $n \sim 1000$
variables seem completely out of the reach of either algorithm.
While formulas in this overconstrained regime
are still challenging, the presence of many forced
steps allows both solvers to completely explore the space fairly
quickly.

\begin{figure}
\begin{center}
\begin{minipage}{3in}
\begin{center}
        \centerline{\hbox{
        \psfig{figure=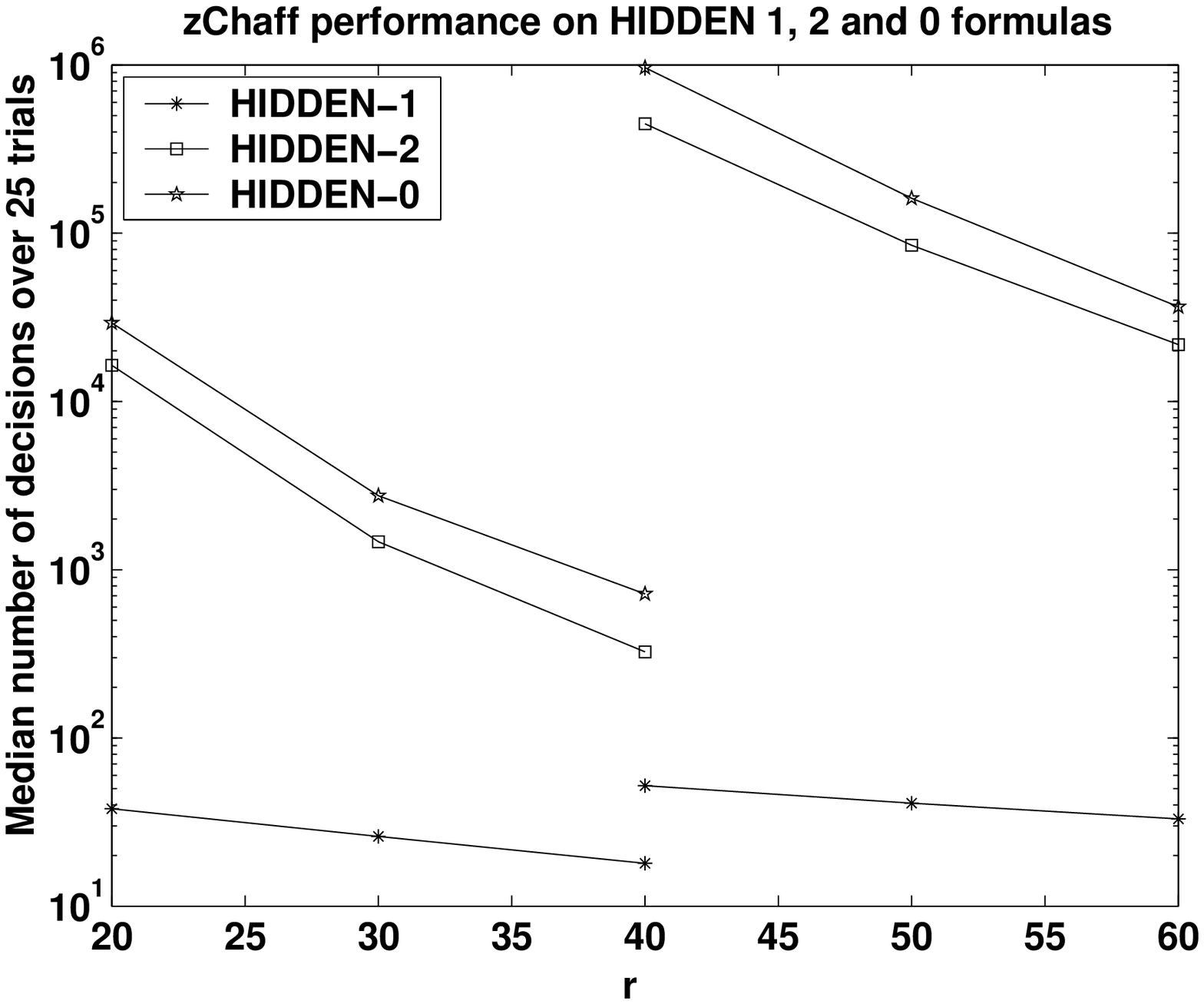,width=3in}
        }}
\end{center}
\end{minipage}\    \
\begin{minipage}{3in}
\begin{center}
        \centerline{\hbox{
        \psfig{figure=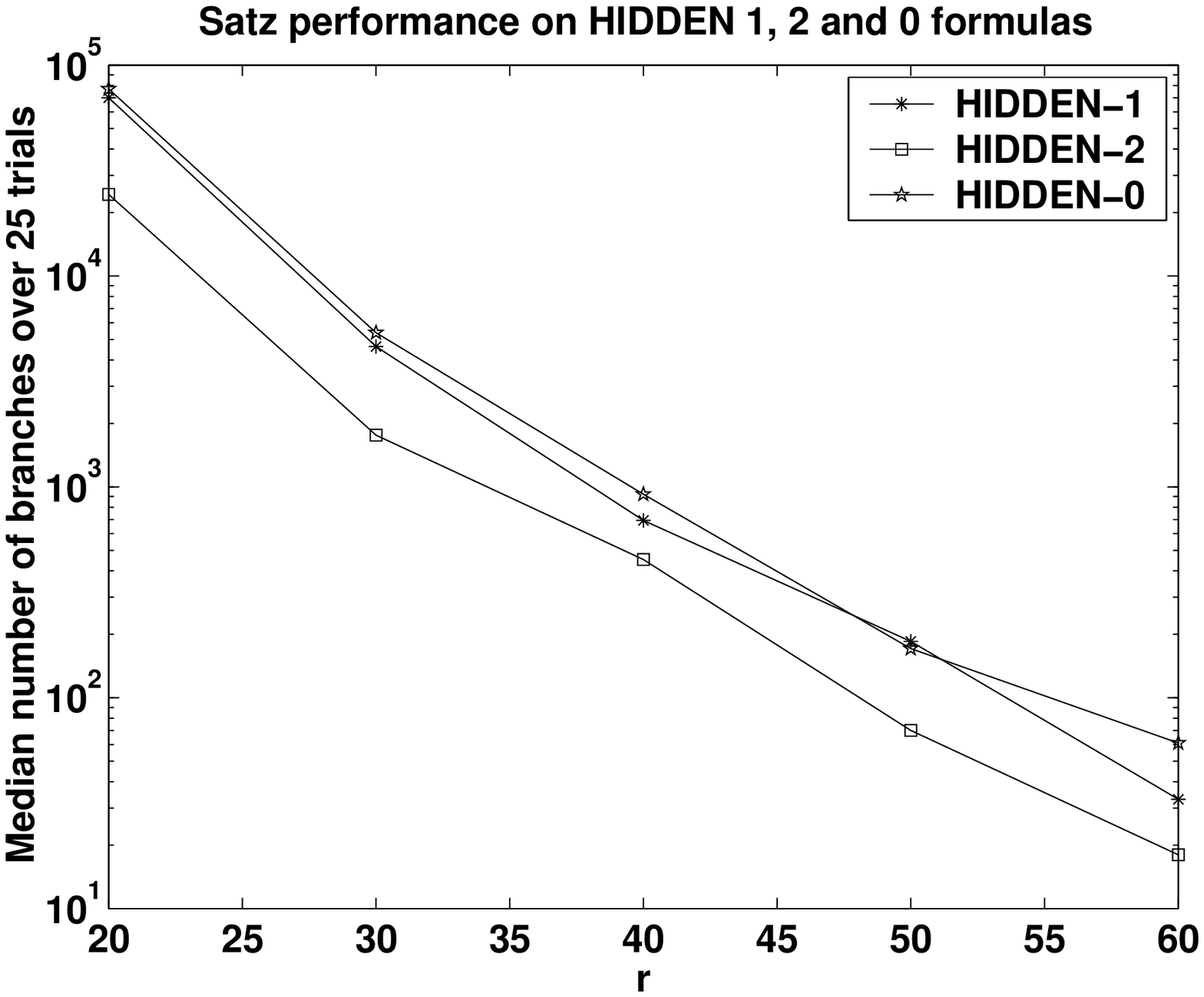,width=3in}
        }}
\end{center}
\end{minipage}
\end{center}
\caption{The median number of branchings made by \zchaff\ and
\satz\  on random instances with 0, 1, and 2 hidden assignments
(on a $\log_{10}$ scale).  For \zchaff\ we use $n = 1000$ for $r =
20, 30, 40$ and $n = 3000$ for $r = 40, 50, 60$, and for \satz\ we
use $n = 3000$ throughout.  Each point is the median of 25 trials.
The 2-hidden formulas are almost as hard for both algorithms as
the 0-hidden ones, while the 1-hidden formulas are much easier for
\zchaff.} \label{fig:zchaffsatz}
\end{figure}

We obtained \zchaff\ from the Princeton web
site~\cite{zchaffsite}. The left part of Figure~\ref{fig:zchaffsatz} shows its
performance on random formulas of all three types (with $n=1000$
for $20 \le r \le 40$ and $n = 3000$ for $40 \le n \le 60$).  We
see that the number of decisions for all three types of problems
decreases rapidly as $r$ increases, consistent with earlier
findings for complete solvers on random 3-SAT formulas.

Figure~\ref{fig:zchaffsatz} shows that \zchaff\ finds 2-hidden
formulas almost as difficult as 0-hidden ones, which for this
range of $r$ are unsatisfiable with overwhelming probability.  On
the other hand, the 1-hidden formulas are much easier, with a
number of branchings between 2 and 5 orders of magnitude smaller.
It appears that while \zchaff's smarts allow it to quickly ``zero
in'' on a single hidden assignment, the attractions exerted by a
complementary pair of assignments do indeed cancel out, making
2-hidden formulas almost as hard as unsatisfiable ones. That is,
the algorithm eventually ``stumbles'' upon one of the two hidden
assignments after a search that is nearly as exhaustive as for the
unsatisfiable random 3-SAT formulas of the same density.

We obtained \satz\ from the SATLIB web site \cite{satlib}.  The
right part of Figure~\ref{fig:zchaffsatz} shows experiments on
random formulas of all three types with $n=3000$.  As for \zchaff,
the 2-hidden formulas are almost as hard for \satz\ as the
0-hidden formulas are.  On the other hand, the 1-hidden formulas
are too. Indeed, Figure~\ref{fig:zchaffsatz} shows that the median
number of branches for all three types of formulas is within a
multiplicative constant.

The reason for this is simple: while \satz\ makes intelligent
decisions about which variable to branch on, it tries these
branches in a fixed order, attempting first to set each variable
false~\cite{satzpaper}. Therefore, a single hidden assignment will
appear at a uniformly random leaf in \satz's search tree. In the
2-hidden case, since the two hidden assignments are complementary,
one will appear in a random position and the other one in the
symmetric position with respect to the search tree. Naturally,
trying branches in a fixed order is a good idea when the true goal
is to prove that a formula is unsatisfiable, \eg in hardware
verification. However, we expect that if \satz\ were
modified to, say, use the majority heuristic to choose a
variable's first value, its performance on the three types of
problems would be similar to \zchaff's.

\subsection{\survey}

\survey\ is an incomplete solver recently introduced by M\'{e}zard
and Zecchina~\cite{mz} based on a generalization of belief
propagation the authors call {\em survey propagation}. It is
inspired by the physical notion of ``replica symmetry breaking''
and the observation that for $3.9 < r < 4.25$, random 3-SAT
formulas appear to be satisfiable, but their satisfying
assignments appear to be organized into clumps.

\begin{figure}
\centerline{\psfig{figure=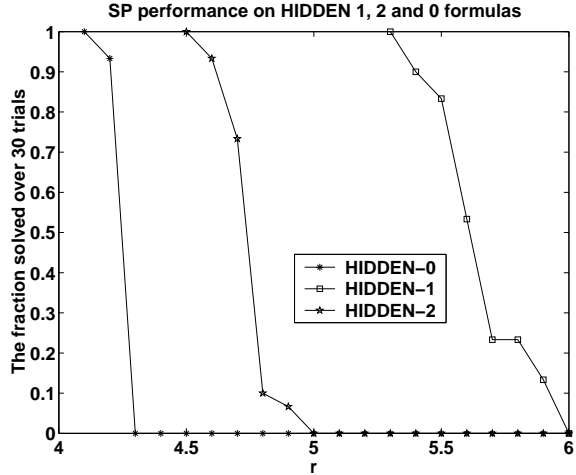,width=3in}}
\caption{The fraction of problems successfully solved by \survey\
as a function of density, with $n=10^4$ and $30$ trials for each
value of $r$.  The threshold for solving 2-hidden formulas is
somewhat higher than for 0-hidden ones, and for 1-hidden formulas
it is higher still.} \label{fig:survey}
\end{figure}

In Figure~\ref{fig:survey} we compare \survey's performance on the
three types of problems near the satisfiability threshold.  For
$n=10^4$ \survey\ solves 2-hidden formulas at densities somewhat
above the threshold, up to $r \approx 4.8$, while it solves the
1-hidden formulas at still higher densities, up to $r \approx
5.6$.

Presumably the 1-hidden formulas are easier for \survey\ since the
``messages'' from clauses to variables, like the majority
heuristic, tend to push the algorithm towards a hidden assignment.
Having two hidden assignments appears to cancel these messages out
to some extent, causing \survey\ to fail at a lower density.
However, this argument does not explain why the \survey-threshold
for 2-hidden formulas should be higher than the satisfiability
threshold; nor does it explain why \survey\ does not solve 1-hidden
formulas for arbitrarily large $r$.  Indeed, we find this latter
result surprising, since as $r$ increases the majority of clauses
should point more and more consistently towards the hidden
assignment in the 1-hidden case.

We note that we also performed the above experiments with $n=2
\times 10^4$ and with $5000$ iterations, instead of the default
$1000$, for \survey's convergence procedure. The thresholds of
Figure~\ref{fig:survey} for 1-hidden and 2-hidden formulas
appeared to be stable under both these changes, suggesting that
they are not merely artifacts of our particular experiments. We
propose investigating these thresholds as a direction for further
work.

\subsection{\walksat}

We conclude with a local search algorithm, \walksat.  Unlike the
complete solvers, \walksat\ can solve problems with $n=10^4$
fairly close to the threshold.
We performed experiments 
both with a random initial state, and with a biased initial state
where the algorithm starts with $75\%$ agreement with one of the
hidden assignments (note that this is exponentially unlikely).
In both cases, we performed trials of $10^8$ flips for each formula, without random restarts,
where each step does a random or greedy flip with equal probability.
Since random initial states \whp\ have roughly $50\%$ agreement with both
hidden assignments, we expect their attractions to cancel out
so that \walksat\ will have difficulty finding either of them.
On the other hand, if we begin with a biased initial state, then the
attraction from the nearby assignment will be much stronger than
the other one; this situation is similar to a 1-hidden formula,
and we expect \walksat\ to find it easily.  Indeed our data confirms these expectations.

\begin{figure}
\begin{center}
\begin{minipage}{3in}
\begin{center}
        \centerline{\hbox{
        \psfig{figure=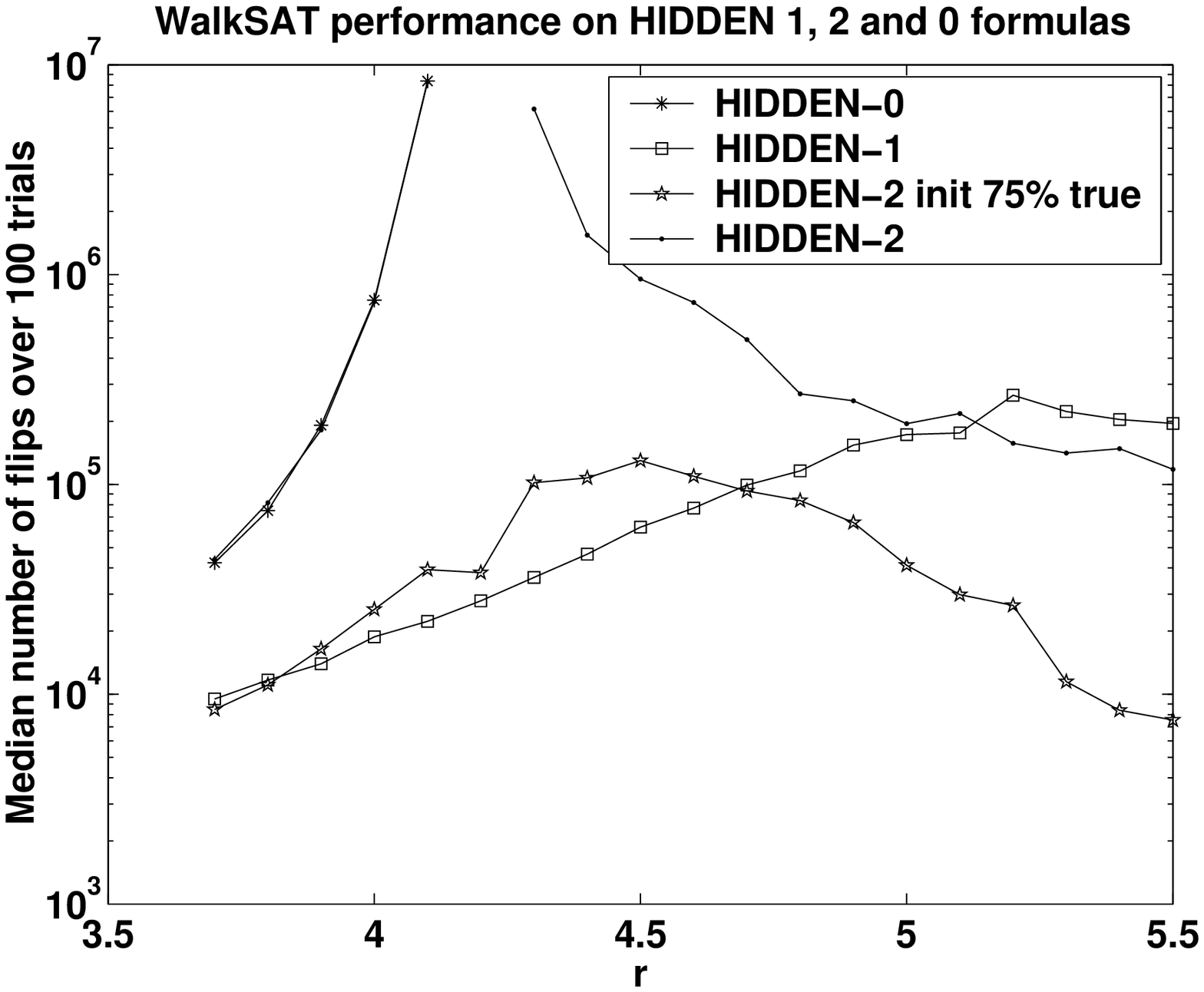,width=3in}
        }}
\end{center}
\end{minipage}\    \
\begin{minipage}{3in}
\begin{center}
        \centerline{\hbox{
        \psfig{figure=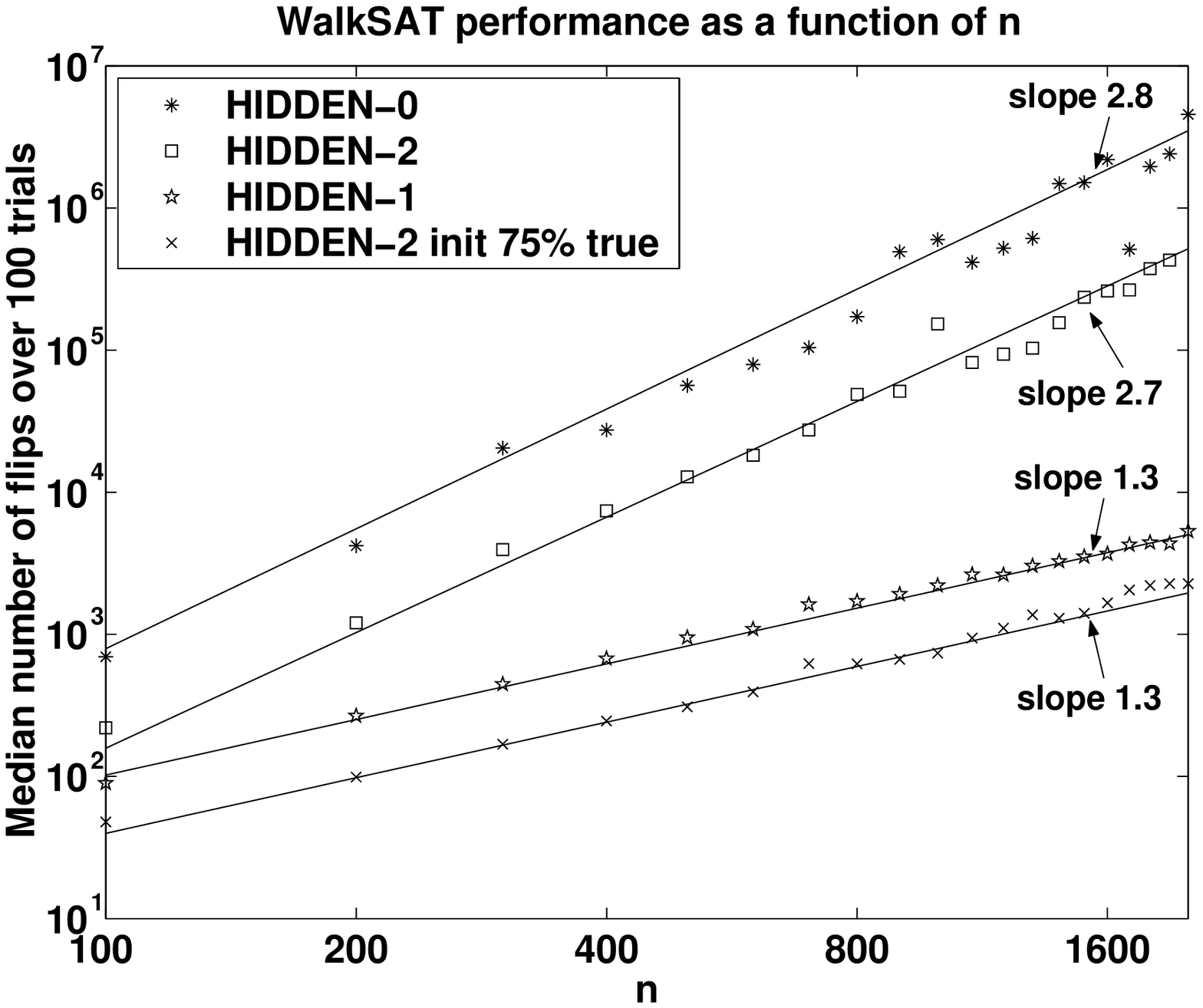,width=3in}
        }}
\end{center}
\end{minipage}
\end{center}
\vspace*{-0.8cm} \caption{Left, the median number of flips needed
by \walksat\ for formulas of all three types below and above the
threshold, with $n = 10^4$. Below the threshold, 2-hidden formulas
are just as hard as 0-hidden ones (they coincide to within the
resolution of the figure) and their running time increases steeply
as we approach the threshold.  Both above and below the threshold,
2-hidden formulas are much harder than 1-hidden ones, unless the
algorithm starts with a (exponentially lucky) biased initial
state. Right, the median number of flips needed by \walksat\ to
solve the three types of formulas at $r=4.25$ as a function of
$n$. Here $n$ ranges from $100$ to $2000$.  While the median
running time for all three is polynomial, the 2-hidden problems
are much harder than the 1-hidden ones unless the have a biased
initial state, with a running time that scales similarly to
0-hidden problems, \ie random 3-SAT.} \label{fig:walksatfig}
\end{figure}

In the first part of~Figure~\ref{fig:walksatfig} we measure
\walksat's performance on the three types of problems with
$n=10^4$ and $r$ ranging from $3.7$ to $5.5$, and compare them
with 0-hidden formulas for $r$ ranging from $3.7$ up to $4.1$, just below
the threshold where they become unsatisfiable.
We see that, below the threshold, the 2-hidden formulas are just as hard
as the 0-hidden ones when \walksat\ sets its initial state randomly;
indeed, their running times coincide to within the
resolution of the figure!  They both become hardest when $r \approx 4.2$,
where $10^8$ flips no longer suffice to solve them.

On the other hand, the 1-hidden formulas are much easier than the 2-hidden
ones, and their running time peaks around $r=5.2$.
Finally, the 2-hidden formulas are
much easier to solve when we start with a biased initial state,
in which case the running time is closer to that of 1-hidden formulas.

In the second part of Figure~\ref{fig:walksatfig}, we compare the three types
of formulas at a density very close to the threshold, $r=4.25$, and measure
their running times as a function of $n$.
The data suggests that 2-hidden formulas with random initial states
are much harder than 1-hidden ones, while 2-hidden formulas with
biased initial states have running times within a constant of that of
1-hidden formulas.  (Note that, consistent with experiments of~\cite{Barthel},
the median running time of all three types of problems is polynomial in $n$.)

Based on this, we conjecture the following.
Our 2-hidden formulas are just as hard for \walksat\ as 0-hidden
ones up to the satisfiability
threshold, and they are hardest at or near the  threshold.
Moreover, they are much harder than 1-hidden formulas,
unless \walksat\ is lucky enough to have an initial state biased
towards one of the hidden assignments.

\section{Conclusions}

We have introduced an extremely simple new generator of random
satisfiable 3-SAT instances which is amenable to all the
mathematical tools developed for the rigorous study of random
3-SAT instances. Experimentally, our generator appears to produce
instances that are as hard as random 3-SAT instances, in sharp
contrast to instances with a single hidden assignment. This
hardness appears quite robust; our experiments have demonstrated
it both above and below the satisfiability threshold, and for
algorithms that use very different strategies, \ie DPLL solvers
(\zchaff\ and \satz), local search algorithms (\walksat), and
survey propagation (\survey).

We believe that random 2-hidden instances could make excellent
satisfiable benchmarks, especially just around the satisfiability
threshold, say at $r = 4.25$ where they appear to
be the hardest for \walksat\ (although beating \survey\ requires somewhat
higher densities).

Several aspects of our experiments suggest exciting directions for further work,
including:
\begin{enumerate}
\item Proving that the expected running time of natural Davis-Putnam algorithms
on 2-hidden formulas is exponential in $n$ for $r$ above some critical density.
\item Explaining the different threshold behaviors of
\survey\ on 1-hidden and 2-hidden formulas.
\item Understanding how long \walksat\ takes at the
midpoint between the two hidden assignments, before it becomes
sufficiently unbalanced to converge to one of them.
\item Studying random 2-hidden formulas in the dense case where there are
$\omega(n)$ clauses.
\end{enumerate}

\bibliographystyle{amsplain}
\bibliography{2hidden}

\end{document}